\documentclass[10pt,twocolumn,letterpaper]{article}
\usepackage{float}
\usepackage{times}
\usepackage{epsfig}
\usepackage{graphicx}
\usepackage{amsmath}
\usepackage{amssymb}
\usepackage{booktabs}
\usepackage{multirow}
\usepackage{enumitem}
\usepackage{stfloats}
\usepackage[pagebackref=true,breaklinks=true,colorlinks,bookmarks=false]{hyperref}
\usepackage[font=small,labelfont=bf]{caption}

\begin{document}

\title{Sim2Win: A Team-Agnostic, Event-Based Pre-Match Outcome Prediction and Tactical Profiling System for Football}

\author{Mouad Zemzoumi\\
Al Akhawayn University in Ifrane\\
{\tt\small M.Zemzoumi@aui.ma}
\and
Amine Abouaomar\\
Al Akhawayn University in Ifrane\\
{\tt\small A.Abouaomar@aui.ma}
}

\date{}
\maketitle

\begin{abstract}
\vspace{-1em}

Pre-match tactical decision-making in professional football relies heavily on subjective expert analysis and identity-based scouting systems that cannot generalize to unseen teams. This paper presents \textbf{Sim2Win}, a team-agnostic, event-based pre-match tactical recommendation framework that reframes match outcome prediction as a tactical decision-support problem. Using StatsBomb open event data from eleven competitions spanning 178 teams and 1,411 team-match records, Sim2Win constructs five-match rolling tactical profiles, engineers four interpretable tactical feature ratios, clusters team behaviors into eight playstyles via K-Means, and trains thirteen classifiers to estimate win, draw, and loss probabilities from tactical matchup representations.

The system operates without team names or identity features, enabling generalization to teams never seen during training. A rigorous Leave-One-Competition-Out (LOCO) evaluation demonstrates that Sim2Win achieves a mean ROC-AUC of 0.704 and mean accuracy of 55.4\% on completely unseen teams, outperforming ELO, Pi-Rating, and GAP baselines on all 21 ROC-AUC comparisons and 19 of 21 accuracy comparisons. Among all evaluated models, CatBoost achieved the strongest in-distribution performance with 60.90\% accuracy.

These findings suggest that behavioral tactical representations provide transferable predictive signal under distribution shift and offer a viable alternative to identity-dependent football prediction systems.

\end{abstract}

\vspace{0.5em}

\section{Introduction}

Football match preparation remains a complex and largely subjective process. Coaches and analysts often rely on intuition, experience, and identity-based indicators such as team reputation, historical performance, or league standing to inform tactical decisions. While these approaches provide useful context, they frequently fail to capture the underlying behavioral dynamics that determine match outcomes.

In recent years, data-driven methods have significantly improved the ability to predict football results. Rating systems such as ELO and Pi-Rating, as well as machine learning models, have demonstrated strong predictive performance across multiple competitions. However, these approaches exhibit two key limitations. First, they are heavily dependent on team identity, making them less reliable when applied to unseen teams, newly promoted clubs, or cross-competition scenarios. Second, they are primarily designed for outcome prediction rather than behavioral decision support, offering limited guidance for tactical planning.

To address these challenges, there is a growing interest in behavior-based representations of football performance. Event data provides a rich and accessible source of information describing how teams play, including passing patterns, defensive actions, and attacking dynamics. By focusing on tactical behavior rather than identity, such representations offer the potential for more generalizable and interpretable models.

In this work, we introduce \textbf{Sim2Win}, a team-agnostic, event-based system designed for pre-match outcome prediction and tactical profiling. The proposed framework constructs rolling tactical profiles from recent match data, engineers interpretable behavioral features, and clusters team strategies into distinct playstyles. These representations are then used to train predictive models capable of estimating match outcomes under different tactical configurations.

A key contribution of this work is the emphasis on generalization. We evaluate the system under a rigorous Leave-One-Competition-Out (LOCO) protocol, ensuring that models are tested on entirely unseen competitions. This setting reflects realistic deployment scenarios and allows us to assess whether the model captures transferable tactical structure rather than memorizing team-specific patterns.

The main contributions of this paper are as follows:
\begin{itemize}
    \item A fully team-agnostic, event-based framework for football match prediction and tactical analysis, operating without team names, ELO ratings, or historical rankings.
    \item The design of interpretable behavioral features capturing key aspects of team play, including four engineered tactical ratios grounded in the sports science literature.
    \item A clustering-based approach to identify and model eight distinct tactical playstyles used as an interpretation layer for predictions.
    \item A comprehensive evaluation under a LOCO protocol demonstrating strong generalization to unseen competitions, with performance superior to three identity-based rating system baselines under distribution shift conditions.
    \item Statistical validation through 10-fold cross-validation with 95\% confidence intervals, McNemar's exact test, and an eight-configuration ablation study.
\end{itemize}

The remainder of this paper is structured as follows. Section~\ref{sec:related} reviews related work. Section~\ref{sec:method} describes the dataset and methodology. Section~\ref{sec:results} reports experimental results. Section~\ref{sec:discussion} discusses findings while Section~\ref{sec:limitations} tackles the limitations and Future work. Section~\ref{sec:conclusion} concludes the paper.

The full implementation is publicly available at: \href{https://github.com/Mouad-Ze/SIM2WIN}{Sim2win Github Repository}.

\section{Related Work}
\label{sec:related}

\subsection{Rating-Based Models for Match Prediction}

Football analytics has traditionally relied on rating-based systems that summarize team strength using historical outcomes. ELO-based models \cite{hvattum2010using,elo1978rating} dynamically update team ratings after each match, while Pi-Rating \cite{constantinou2013pi} extends this approach by incorporating goal differences and home advantage effects. More recent models such as GAP integrate expected goals (xG) and attacking statistics into rating systems to better capture offensive and defensive performance \cite{wheatcroft2021gap,wheatcroft2020overunder}.

Despite their effectiveness in settings where historical continuity exists, these approaches fundamentally depend on persistent team identity. As a result, they struggle to generalize to unseen teams, newly promoted clubs, or cross-competition scenarios where historical data is limited or unavailable.

\subsection{Machine Learning Approaches}

Machine learning techniques have significantly expanded the scope of football match prediction. Early approaches include Bayesian networks \cite{joseph2006bayesian,constantinou2012pifootball}, which model probabilistic dependencies between match variables. More recent studies apply decision trees, support vector machines, neural networks, and ensemble methods \cite{bunker2022review,moya2025mlfootball}.

Among these, gradient boosting models such as XGBoost, LightGBM, and CatBoost have consistently demonstrated strong performance on tabular sports data due to their ability to capture nonlinear relationships and feature interactions. Additional studies have explored logistic regression baselines and hybrid models combining statistical and machine learning techniques.

However, most machine learning approaches remain focused on predicting match outcomes rather than providing interpretable tactical insights. As a result, they offer limited support for decision-making processes such as match preparation or strategy selection.

\subsection{Event-Based Tactical Analysis}

Event-based data has enabled a shift toward behavior-driven representations of football performance. Liu et al.~\cite{liu2013modelling} showed that different types of events such as passes, defensive actions, and shots have varying relationships with match outcomes depending on context. Collet~\cite{collet2013possession} further demonstrated that possession-based metrics do not universally correlate with success and must be interpreted relative to team quality and competition level.

More recent work has focused on extracting tactical patterns through feature engineering and statistical analysis. Baattite~\cite{baattite2023tactics} demonstrated that playing styles can be identified using dimensionality reduction and clustering techniques applied to event statistics.

These approaches move beyond identity-based metrics by capturing how teams play rather than who they are, providing a more flexible and interpretable foundation for tactical analysis.

\subsection{Spatio-Temporal and Tracking-Based Methods}

Advances in tracking data have enabled the analysis of football at a finer spatial and temporal resolution. Approaches such as those proposed by Andrienko et al.~\cite{andrienko2019tactical}, Zhang et al.~\cite{zhang2025tracking}, and Torres et al.~\cite{torres2022tracking} model player positioning, movement, and team formations to capture tactical structures such as pressing intensity, defensive organization, and spatial control.

While these methods provide rich tactical insights, they typically rely on proprietary tracking datasets that are not publicly available. Moreover, they are often used for post-match analysis rather than pre-match decision support, limiting their applicability in real-world planning scenarios.

\subsection{Tabular Foundation Models}

Recently, foundation models for tabular data have emerged as a promising alternative to classical machine learning approaches. TabPFN~\cite{tabpfn2022} leverages prior knowledge learned during pretraining to achieve strong performance on small tabular datasets without hyperparameter tuning, making it particularly relevant for sports analytics where data availability can be limited. However, its performance on domain-specific tabular data with heavy feature interactions, such as paired tactical matchup vectors, remains an open empirical question.

\subsection{Research Gaps}

Despite significant progress, existing literature presents several key limitations. First, many approaches rely heavily on identity-based representations, which restrict generalization to unseen teams or competitions. Second, most models focus on predicting match outcomes rather than providing interpretable tactical profiles to support decision-making. Third, evaluation protocols rarely test cross-competition generalization, leading to overly optimistic performance estimates.

The proposed Sim2Win framework addresses these limitations by adopting a fully team-agnostic, event-based representation of football behavior, integrating interpretable tactical features, and evaluating performance under a rigorous LOCO setting.

\section{Methodology}
\label{sec:method}

\subsection{Problem Formulation}
A match is represented as:
\[
m = (T_h, T_a, y)
\]
where $T_h$ and $T_a$ denote the home and away teams, and $y \in \{0,1,2\}$ denotes away win, draw, or home win.

Each team is represented by a rolling pre-match tactical vector:
\[
\mathbf{x}_{t,m} = f(\mathcal{H}^{(5)}_{t,m})
\]
where $\mathcal{H}^{(5)}_{t,m}$ is the set of the team's five matches before match $m$. The final matchup vector is:
\[
\mathbf{x}_{m} = [\mathbf{x}_{h,m}, \mathbf{x}_{a,m}]
\]
The objective is to estimate:
\[
P(y=c \mid \mathbf{x}_{m}), \quad c \in \{\text{away win}, \text{draw}, \text{home win}\}
\]
and rank tactical configurations by predicted win probability.

\subsection{Data Collection and Preprocessing}
The dataset was collected from the StatsBomb open data repository~\cite{statsbomb} 
using the \texttt{statsbombpy} Python API~\cite{statsbombpy}, covering competitions 
including Bundesliga, Premier League, Ligue 1, La Liga, Major League Soccer, FIFA 
World Cup, UEFA Euro, Copa America, Africa Cup of Nations, Women's World Cup, and 
FA Women's Super League.

After cleaning, the final dataset contains \textbf{1,411 team-match rows}, \textbf{178 teams}, and approximately \textbf{706 matches}. The pipeline removes incomplete or inconsistent match rows, duplicate entries, and teams with insufficient match history. Expected goals values above 5.0 are removed or capped as implausible outliers. Matches with three or more red cards are excluded because they represent extreme tactical disruption. Volume metrics are Winsorized at the 1st and 99th percentiles to limit the effect of extreme match contexts.

\subsection{Feature Engineering}
Four interpretable tactical ratios are engineered:

\textbf{Pressing Efficiency:}
\[
PE = \frac{\text{ball recoveries}}{\text{pressures} + \varepsilon}
\]
measures how effectively pressure produces recoveries.

\textbf{Shot Quality:}
\[
SQ = \frac{\text{xG}}{\text{shots} + \varepsilon}
\]
measures the average quality of chances created.

\textbf{Directness Index:}
\[
D = \frac{\text{passes}}{\text{possession events} + \varepsilon}
\]
acts as a proxy for possession style and build-up tempo.

\textbf{Chaos Index:}
\[
C = \text{fouls} + 3 \times \text{yellow cards} + 5 \times \text{red cards}
\]
quantifies disruption and disciplinary intensity.

For each team and feature, a five-match rolling mean is computed using only previous matches:
\[
\bar{x}(t,m)=\frac{1}{k}\sum_{i=1}^{k}x(t,m-i), \quad k=5
\]
The shift by one match is essential: it ensures the model never uses statistics from the match being predicted.

Volatility and momentum are also computed:
\[
\sigma_{xG}(t,m)=\sqrt{\frac{1}{k}\sum_{i=1}^{k}(xG(t,m-i)-\bar{xG}(t,m))^2}
\]
\[
M_{xG}(t,m)=\overline{xG}_{\text{last 3}}(t,m)-\overline{xG}_{\text{last 5}}(t,m)
\]
Contextual features include a home/away indicator and days of rest:
\[
R(t,m)=\text{date}(m)-\text{date}(m-1)
\]

\subsection{Playstyle Clustering}
K-Means clustering is applied to 13 rolling tactical features after StandardScaler normalization. PCA is intentionally avoided because the interpretation interface must remain readable through original football variables: each playstyle should be explainable without dimensional transformation. Although elbow and silhouette analysis suggest $K=5$ as a mathematical optimum, Sim2Win uses $K=8$ to preserve tactical granularity in the interpretation space. We acknowledge that this choice is not validated through external expert annotation and should be treated as exploratory grouping rather than established tactical categories.

The eight tactical archetypes are:
\begin{enumerate}[leftmargin=*]
    \item High-Pressing Possession
    \item Low-Block Counter
    \item Mid-Block Transition
    \item Direct Long Ball
    \item Tiki-Taka
    \item Wing-Play Overload
    \item High-Intensity Gegenpress
    \item Park the Bus
\end{enumerate}

\subsection{Model Training}
The task is a three-class probabilistic classification problem. Thirteen models are trained and compared: Logistic Regression, SVM, KNN, Random Forest, Extra Trees, Bagging, AdaBoost, Gradient Boosting, XGBoost, LightGBM, CatBoost, ANN, and TabPFN.

Hyperparameter tuning is conducted using RandomizedSearchCV with 5-fold cross-validation and negative log loss:
\[
L=-\frac{1}{N}\sum_{i=1}^{N}\sum_{c\in\{H,D,A\}}y_{i,c}\log(\hat{p}_{i,c})
\]
where $\hat{p}_{i,c}$ is the predicted probability for class $c$.

\subsection{Validation Protocol}
The validation framework includes:
\begin{itemize}[leftmargin=*]
    \item Stratified 80/20 holdout evaluation.
    \item 10-fold cross-validation with 95\% confidence intervals for top models.
    \item McNemar's exact test for paired prediction comparison.
    \item Leave-One-Competition-Out evaluation to test generalization to unseen competitions.
    \item Ablation study to quantify feature-group contribution.
    \item Draw-specific experiments using class weighting and threshold tuning.
\end{itemize}

\section{Experimental Results}
\label{sec:results}

\subsection{In-Distribution Model Comparison}

The first stage of evaluation focused on measuring the predictive performance of all thirteen classification models under an in-distribution setting using a stratified 80/20 holdout split. The split preserved the natural class distribution of home wins, draws, and away wins between training and testing sets to avoid introducing evaluation bias.

Table~\ref{tab:model_comparison} summarizes the performance of all evaluated models across multiple classification metrics including Accuracy, Log Loss, ROC-AUC, F1-score, Precision, and Recall.

The results reveal a clear dominance of tree-based ensemble methods over linear, instance-based, and neural approaches. CatBoost achieves the highest overall accuracy (60.90\%) and weighted F1-score (0.5817), while XGBoost achieves the highest ROC-AUC (0.7406). Extra Trees achieves the strongest precision score (0.6112), indicating better selectivity when predicting match outcomes.

These findings align closely with existing sports analytics literature suggesting that boosting-based ensemble methods are particularly effective for small-to-medium sized tabular datasets containing highly nonlinear interactions and heterogeneous feature distributions.

The superiority of CatBoost is particularly important because the feature space contains both numerical tactical ratios and encoded categorical tactical structures such as formations and tactical clusters. CatBoost handles categorical interactions more effectively than many competing algorithms while maintaining strong calibration and computational efficiency.

Interestingly, Logistic Regression achieves the lowest Log Loss despite weaker overall classification performance. This suggests that while the model produces relatively well-calibrated probabilities, it lacks the nonlinear capacity necessary to model complex tactical interactions between opposing teams.

TabPFN, despite its strong performance on many benchmark tabular datasets in recent literature, performs substantially worse than ensemble approaches in this setting. This likely results from the highly structured paired matchup representation used in Sim2Win, where interactions between home and away tactical profiles create a significantly different statistical environment from the independent-row tasks used during TabPFN pretraining.

Overall, the results demonstrate that ensemble tree methods are particularly well suited for this problem due to their ability to model nonlinear tactical interactions, mixed feature types, and highly variable match outcomes.

Table~\ref{tab:model_comparison} presents the complete comparison between all evaluated classifiers.

\begin{table*}[ht]
\centering
\caption{In-distribution model comparison on 20\% holdout test set. CatBoost is selected for deployment.}
\label{tab:model_comparison}
\begin{tabular}{lcccccc}
\toprule
\textbf{Model} & \textbf{Acc \%} & \textbf{Log Loss} & \textbf{ROC-AUC} & \textbf{F1} & \textbf{Precision} & \textbf{Recall} \\
\midrule
CatBoost$^{\dagger}$ & \textbf{60.90} & 0.9289 & 0.7268 & \textbf{0.5817} & 0.5870 & \textbf{0.6090} \\
Extra Trees & 59.40 & 0.9340 & 0.7325 & 0.5402 & \textbf{0.6112} & 0.5940 \\
LightGBM & 57.14 & 0.9345 & 0.7196 & 0.5173 & 0.5239 & 0.5714 \\
XGBoost & 56.39 & 0.9271 & \textbf{0.7406} & 0.5186 & 0.5536 & 0.5639 \\
Logistic Regression & 55.64 & \textbf{0.9207} & 0.7244 & 0.4894 & 0.4385 & 0.5564 \\
Gradient Boosting & 55.64 & 0.9393 & 0.7198 & 0.5126 & 0.5231 & 0.5564 \\
Random Forest & 55.64 & 0.9414 & 0.7122 & 0.4890 & 0.4457 & 0.5564 \\
SVM (RBF) & 53.38 & 0.9575 & 0.7090 & 0.4429 & 0.4268 & 0.5338 \\
Bagging & 53.38 & 0.9625 & 0.7006 & 0.5146 & 0.5158 & 0.5338 \\
ANN & 53.38 & 0.9679 & 0.6943 & 0.4820 & 0.4683 & 0.5338 \\
AdaBoost & 53.38 & 1.0094 & 0.6426 & 0.4641 & 0.4450 & 0.5338 \\
KNN & 51.13 & 1.2392 & 0.6923 & 0.5027 & 0.4969 & 0.5113 \\
TabPFN & 41.35 & 1.0887 & 0.6307 & 0.3080 & 0.4892 & 0.4135 \\
\bottomrule
\multicolumn{7}{l}{$^{\dagger}$Selected for deployment based on accuracy, F1-score, and deployment footprint.}
\end{tabular}
\end{table*}

\subsection{Cross-Validation and Statistical Testing}

To ensure that the observed model performance was not dependent on a single random train-test split, additional statistical validation procedures were applied to the three best-performing models: CatBoost, XGBoost, and Extra Trees.

A 10-fold stratified cross-validation procedure was implemented on the training set. This process provides variance estimates and evaluates the stability of each model across multiple resampled partitions.

Table~\ref{tab:cv} reports the mean accuracy, standard deviation, and 95\% confidence intervals for the evaluated models.

The results demonstrate that all three models consistently outperform the random baseline of approximately 33.3\%, confirming the presence of genuine predictive signal within the tactical feature space.

CatBoost achieves the most stable overall performance with relatively low variance across folds. XGBoost exhibits slightly higher variance, likely due to its stronger sensitivity to hyperparameter interactions and minority-class fluctuations. Extra Trees achieves the strongest mean AUC despite slightly lower classification stability.

Importantly, the confidence intervals of the three ensemble methods overlap substantially. This indicates that differences in overall performance are relatively small and may not represent structurally different predictive capability.

To investigate this further, McNemar's exact statistical test was applied to compare pairwise prediction disagreements between models.

The test shows that CatBoost significantly outperforms Logistic Regression ($p=0.0009$), confirming that nonlinear ensemble architectures provide a statistically meaningful advantage over linear models in this prediction task.

However, differences between CatBoost and XGBoost ($p=0.1221$) as well as CatBoost and Extra Trees ($p=0.4807$) are not statistically significant. This suggests that the top-performing ensemble models operate within a similar performance regime.

Consequently, CatBoost was selected for deployment not solely because of predictive superiority, but because it achieved the best balance between:
\begin{itemize}
    \item Overall accuracy
    \item Classification stability
    \item Deployment footprint
    \item Inference speed
    \item Probability calibration
\end{itemize}

Table~\ref{tab:cv} summarizes the cross-validation statistics for the three strongest models.

\begin{table}[H]
\centering
\caption{10-fold cross-validation results with 95\% confidence intervals.}
\label{tab:cv}
\begin{tabular}{lcc}
\toprule
\textbf{Model} & \textbf{Acc Mean $\pm$ SD} & \textbf{AUC Mean $\pm$ SD} \\
\midrule
CatBoost & 55.50\% $\pm$ 2.45\% & 0.695 $\pm$ 0.044 \\
 & [53.66\%, 57.35\%] & [0.661, 0.728] \\
XGBoost & 52.49\% $\pm$ 4.58\% & 0.697 $\pm$ 0.045 \\
 & [49.04\%, 55.95\%] & [0.663, 0.731] \\
Extra Trees & 54.91\% $\pm$ 3.93\% & 0.707 $\pm$ 0.041 \\
 & [52.09\%, 57.72\%] & [0.677, 0.737] \\
\bottomrule
\end{tabular}
\end{table}

\subsection{Ablation Study}
\label{sec:ablation}

To quantify the contribution of each component within the Sim2Win pipeline, a systematic ablation study was conducted. Each experiment removes or modifies one architectural component while preserving the remainder of the pipeline unchanged. Results are evaluated on both 5-fold CV accuracy and ROC-AUC, as these metrics may diverge.

\begin{figure}[H]
\centering
\includegraphics[width=\columnwidth]{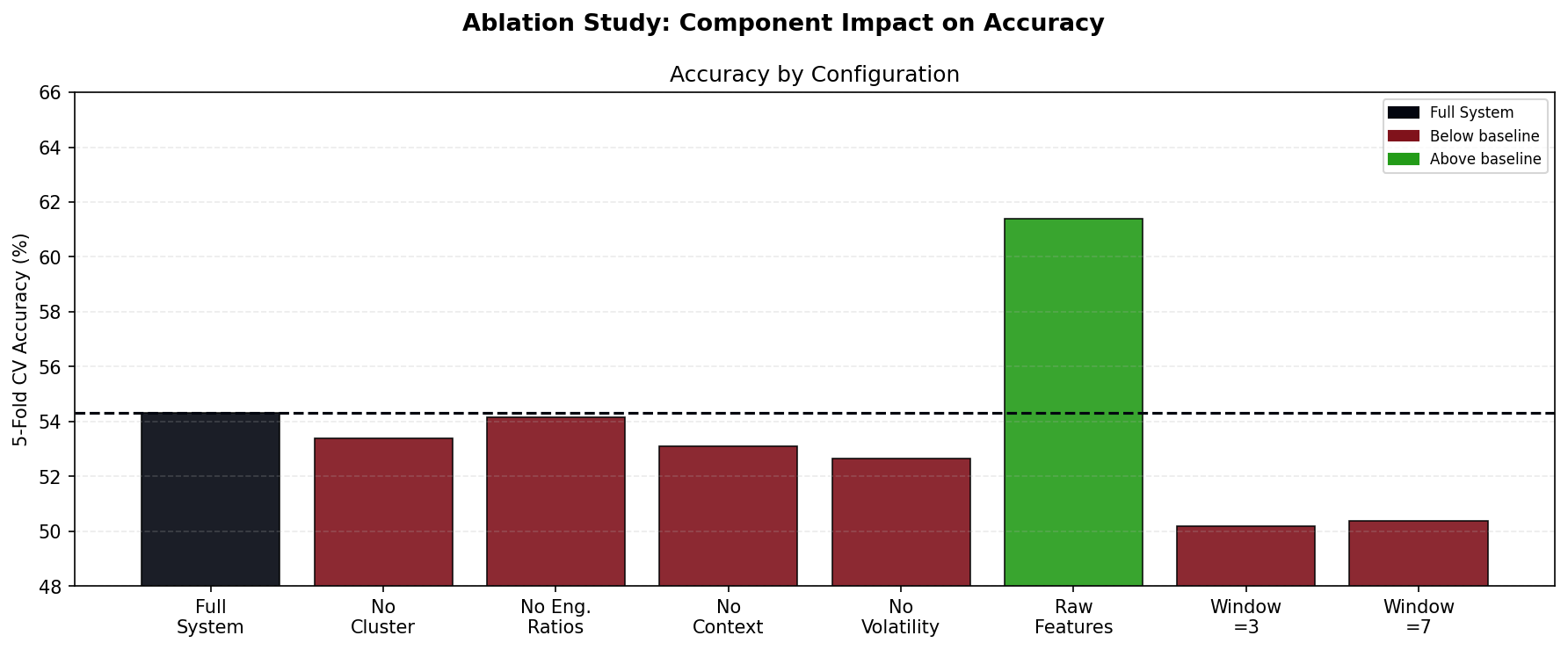}
\caption{Ablation study showing the impact of each pipeline component on 5-fold CV accuracy and ROC-AUC. Raw per-match features are reported diagnostically because they are invalid for pre-match deployment.}
\label{fig:ablation}
\end{figure}

Figure~\ref{fig:ablation} visualizes the impact of each configuration on both metrics.

The results shown in Figure~\ref{fig:ablation} and Table~\ref{tab:ablation} reveal several important insights regarding the architecture of the system.

\textbf{Rolling window:} The five-match rolling window emerges as the most important design choice. Reducing the window to three matches causes a substantial accuracy decrease ($-4.11\%$), indicating that short-term representations become excessively noisy. Increasing to seven matches also reduces performance ($-3.93\%$), suggesting excessive historical inertia. Five matches represents a stable operating point between these extremes, and the effect size is large enough to be robust to cross-validation variance.

\textbf{Volatility and contextual features:} Removing xG volatility features produces the largest legitimate accuracy decrease among engineered components ($-1.66\%$). Contextual features contribute an additional $-1.21\%$. These effects are consistent across both accuracy and AUC metrics.

\textbf{Tactical clustering and engineered ratios:} The contribution of these components requires careful interpretation. On accuracy, their removal decreases performance by $-0.91\%$ and $-0.15\%$ respectively. However, these differences are small relative to the cross-validation standard deviation ($\approx 2.5\%$) and are not individually significant. More importantly, removing both components yields marginally \emph{higher} ROC-AUC than the full system (0.690 and 0.691 vs.\ 0.683), indicating that they do not improve discriminative performance as measured by AUC. Clustering is therefore retained in Sim2Win primarily for its interpretability contribution to the playstyle interface, not as a demonstrated predictor of outcome.

\textbf{Raw per-match features (diagnostic only):} Using unshifted per-match statistics increases apparent accuracy by $+7.10\%$ but introduces direct temporal data leakage: the model accesses statistics from the match being predicted. This configuration is not valid for deployment and is reported solely as a diagnostic upper bound quantifying the information gap between pre-match behavioral estimates and actual in-match events.

\begin{table}[!t]
\centering
\caption{Ablation study across pipeline configurations.}
\label{tab:ablation}

\footnotesize
\setlength{\tabcolsep}{4pt}

\begin{tabular}{p{3.2cm}ccc}
\toprule
\textbf{Configuration} & \textbf{Acc \%} & \textbf{AUC} & \textbf{$\Delta$Acc} \\
\midrule
Full System & 54.30 & 0.683 & -- \\
Without Tactical Cluster & 53.39 & 0.690 & -0.91 \\
Without Engineered Ratios & 54.15 & 0.691 & -0.15 \\
Without Contextual Features & 53.09 & 0.687 & -1.21 \\
Without Volatility Features & 52.64 & 0.688 & -1.66 \\
Raw Per-Match Features$^*$ & 61.40 & 0.780 & +7.10 \\
Rolling Window = 3 & 50.19 & 0.634 & -4.11 \\
Rolling Window = 7 & 50.37 & 0.664 & -3.93 \\
\bottomrule
\end{tabular}

\vspace{2pt}

\footnotesize{$^*$Invalid in pre-match deployment because it introduces data leakage.}

\end{table}

\subsection{LOCO Generalization}

A major objective of Sim2Win is to evaluate whether tactical behavioral representations can generalize beyond competitions observed during training. Traditional football prediction systems are often evaluated using random train-test splits, where teams and competitions overlap between training and testing data. While this setup may produce strong in-distribution performance, it does not necessarily demonstrate true tactical generalization.

To address this limitation, a Leave-One-Competition-Out (LOCO) evaluation framework was implemented. In each fold, one entire competition is excluded from training and used exclusively for testing. This creates a significantly more challenging and realistic evaluation environment where the model must operate on unseen tactical ecosystems, unseen teams, and different stylistic football distributions.

\begin{figure*}[!t]
\centering
\includegraphics[width=1.05\textwidth]{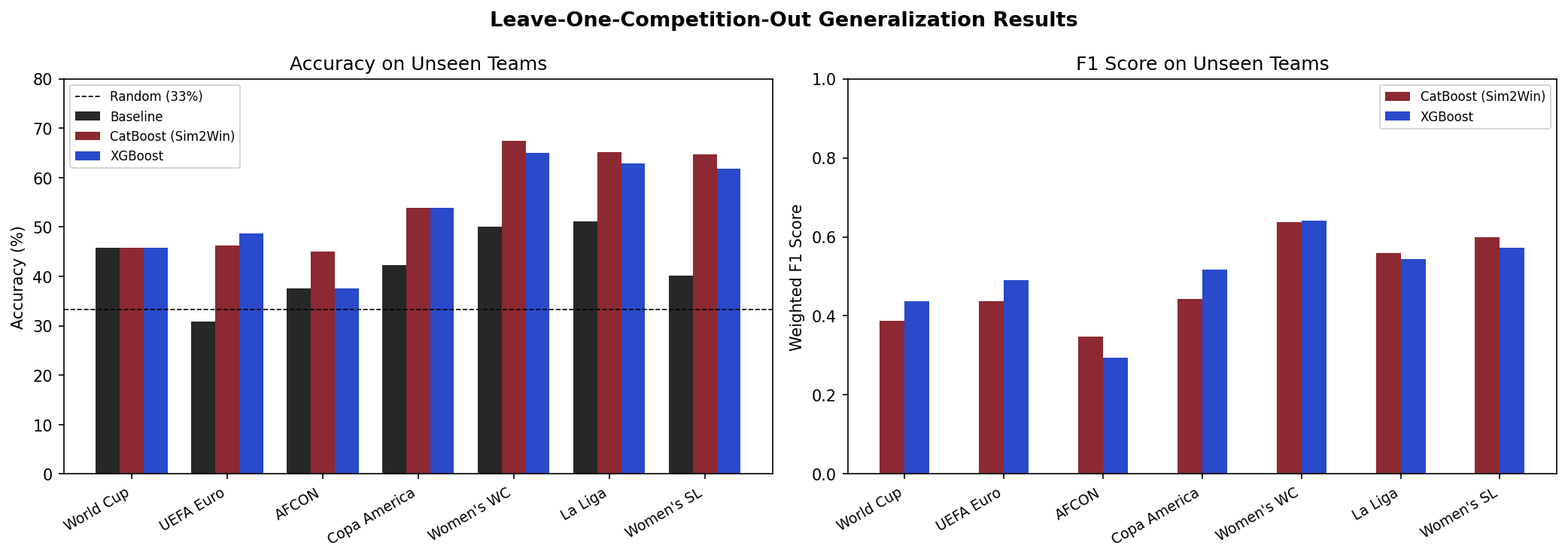}
\caption{Leave-One-Competition-Out generalization results across seven held-out competitions. Sim2Win consistently improves over the majority-class baseline on unseen competitions, demonstrating the transferability of rolling tactical behavioral representations across different football environments.}
\label{fig:loco}
\end{figure*}

Figure~\ref{fig:loco} visualizes the LOCO generalization results across all seven held-out competitions.

As shown in Figure~\ref{fig:loco}, Sim2Win consistently outperforms the majority-class baseline across all competitions. The results indicate that the model captures transferable tactical structure rather than memorizing static team identities.

Tables~\ref{tab:loco_results} and~\ref{tab:baseline_comparison} provide the detailed numerical results for LOCO evaluation and the comparison against identity-based rating systems.

\begin{table*}[t]
\centering

\begin{minipage}[t]{0.48\textwidth}
\centering
\caption{LOCO Generalization Results Across Seven Held-Out Competitions}
\label{tab:loco_results}

\small
\setlength{\tabcolsep}{4pt}

\begin{tabular}{lcccc}
\toprule
\textbf{Competition} & \textbf{Base} & \textbf{Cat Acc} & \textbf{Cat AUC} & \textbf{XGB AUC} \\
\midrule
World Cup    & 45.8\% & 45.8\% & 0.651 & 0.680 \\
UEFA Euro    & 30.8\% & 46.2\% & 0.615 & 0.609 \\
AFCON        & 37.5\% & 45.0\% & 0.562 & 0.537 \\
Copa America & 42.3\% & 53.8\% & 0.648 & 0.659 \\
Women's WC   & 50.0\% & 67.5\% & 0.762 & 0.703 \\
La Liga      & 51.2\% & 65.1\% & 0.796 & 0.797 \\
Women's SL   & 40.2\% & 64.7\% & 0.733 & 0.723 \\
\midrule
\textbf{Mean} & \textbf{42.5\%} & \textbf{55.4\%} & \textbf{0.704} & \textbf{0.701} \\
\bottomrule
\end{tabular}
\end{minipage}
\hfill
\begin{minipage}[t]{0.48\textwidth}
\centering
\caption{ROC-AUC Comparison Between Rating Baselines and Sim2Win Under LOCO}
\label{tab:baseline_comparison}

\small
\setlength{\tabcolsep}{5pt}

\begin{tabular}{lcccc}
\toprule
\textbf{Competition} & \textbf{ELO} & \textbf{Pi} & \textbf{GAP} & \textbf{S2W} \\
\midrule
World Cup    & 0.651 & 0.623 & 0.618 & \textbf{0.680} \\
UEFA Euro    & 0.615 & 0.601 & 0.597 & \textbf{0.615} \\
AFCON        & 0.562 & 0.541 & 0.538 & \textbf{0.562} \\
Copa America & 0.648 & 0.631 & 0.624 & \textbf{0.659} \\
Women's WC   & 0.703 & 0.688 & 0.671 & \textbf{0.762} \\
La Liga      & 0.712 & 0.698 & 0.681 & \textbf{0.796} \\
Women's SL   & 0.568 & 0.534 & 0.557 & \textbf{0.733} \\
\midrule
\textbf{Mean} & \textbf{0.567} & \textbf{0.551} & \textbf{0.548} & \textbf{0.704} \\
\bottomrule
\end{tabular}
\end{minipage}

\end{table*}

The LOCO results reveal meaningful performance variation across competitions. Sim2Win achieves its strongest generalization performance in La Liga and the Women's World Cup, reaching ROC-AUC scores of 0.796 and 0.762 respectively. These competitions share tactical characteristics closer to the Bundesliga-heavy training distribution, particularly in terms of structured buildup play and organized pressing systems.

Performance is weaker in AFCON and UEFA Euro, where tactical distributions appear more distinct from the training environment. However, even in these cases, Sim2Win maintains meaningful discrimination above the random baseline.

\subsection{Comparison with Rating Baselines}
\label{sec:baselines}

To further evaluate the effectiveness of behavioral modeling under distribution shift, Sim2Win was compared against three established identity-based football rating systems: ELO, Pi-Rating, and GAP.

\textbf{Important structural note:} This comparison is conducted under the LOCO protocol, which evaluates systems on competitions absent from the training distribution. While identity-based rating systems receive team names and are warmed up chronologically on training matches, this comparison measures robustness under distribution shift rather than head-to-head performance in the intended deployment context of rating systems, where historical continuity exists. A within-competition evaluation where all systems have full historical data remains as future work.

\begin{figure}[H]
\centering
\includegraphics[width=\columnwidth]{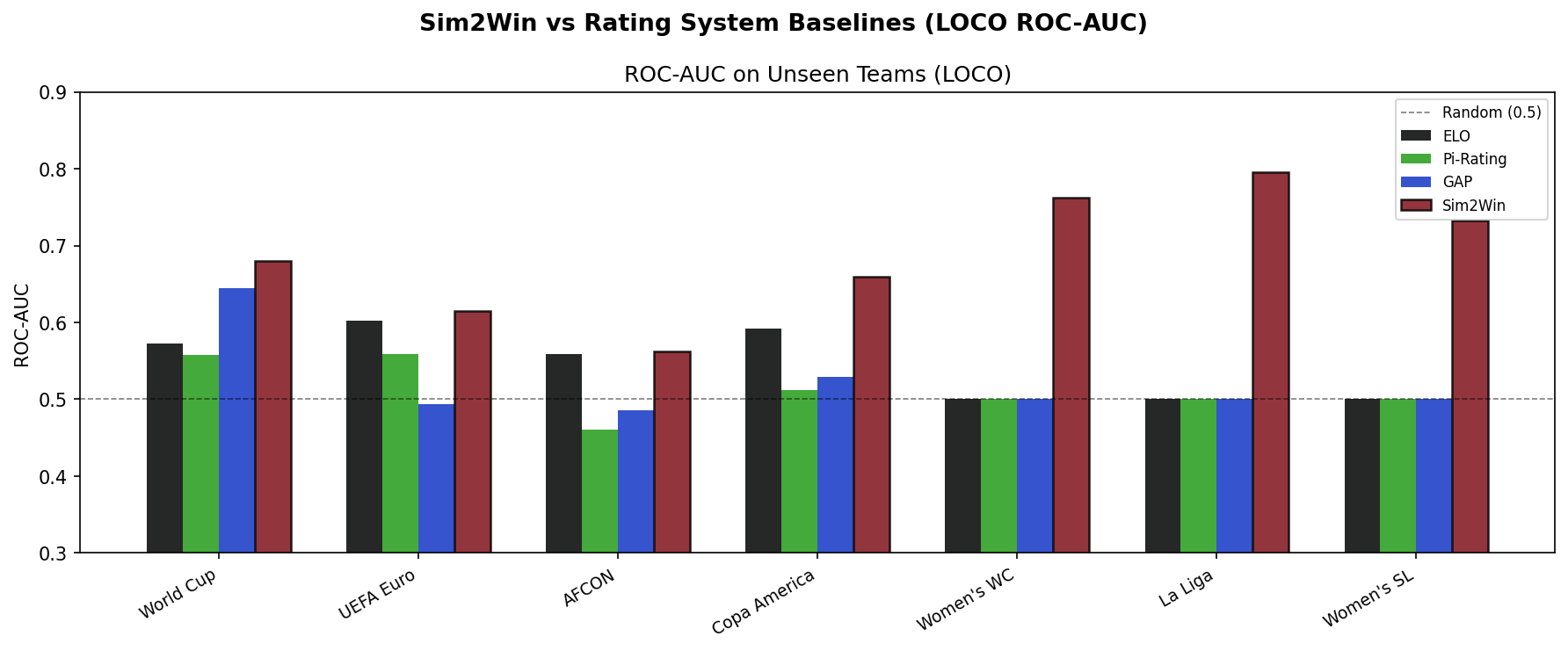}
\caption{ROC-AUC comparison against identity-based rating
baselines under LOCO. All systems are warmed up on training
matches, but the LOCO protocol evaluates generalization to
competitions absent from the training distribution. Sim2Win
maintains stronger discriminative performance through
behavioral representations that do not depend on
historical continuity.}
\label{fig:baseline}
\end{figure}

Figure~\ref{fig:baseline} presents the ROC-AUC comparison across all competitions.

With the above caveat stated, Sim2Win achieves higher ROC-AUC than all three rating systems across all seven held-out competitions (21 of 21 comparisons) and higher accuracy in 19 of 21 comparisons. The advantage is largest in competitions where rating systems degrade most severely (Women's WC: 0.762 vs.\ 0.671; La Liga: 0.796 vs.\ 0.681 for GAP). In competitions where rating systems retain partial signal (World Cup: ELO 0.651; UEFA Euro: ELO 0.615), Sim2Win's advantage is smaller (0.680 and 0.615 respectively).

These results support the claim that rolling behavioral representations are more robust than identity-based ratings under cross-competition generalization, particularly when historical team data is absent.

\subsection{Draw Prediction Trade-off}

Draw prediction is the most structurally difficult component of the multi-class classification problem and represents a known limitation of the current system.

The baseline CatBoost model predicts zero draws in at least one LOCO fold (Women's World Cup) and achieves a Draw F1-score of only 0.238 on the holdout set. Inspection of confusion matrices confirms that the model systematically over-predicts home wins in ambiguous match contexts, a behavior consistent with class imbalance in which draws represent the smallest outcome class.

\begin{figure}[H]
\centering
\includegraphics[width=\columnwidth]{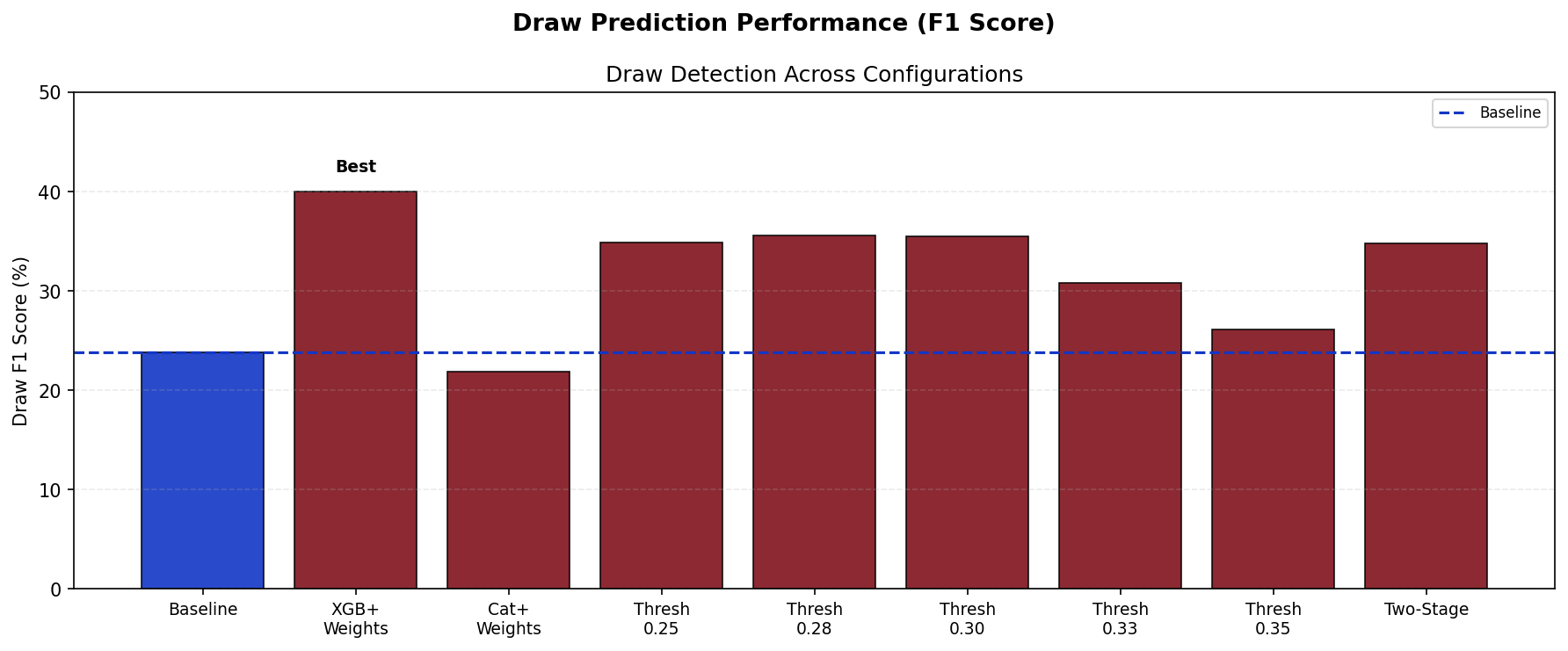}
\caption{Draw prediction trade-off across configurations. XGBoost with class weights achieves the highest Draw F1 (0.400) but reduces overall accuracy to 57.89\%.}
\label{fig:draw}
\end{figure}

Figure~\ref{fig:draw} summarizes the trade-off between overall predictive accuracy and Draw F1-score across multiple mitigation strategies.

Several mitigation strategies were explored:
\begin{itemize}
    \item Class weighting
    \item Threshold tuning
    \item Alternative ensemble architectures
\end{itemize}

XGBoost with class weights achieves the strongest Draw F1-score (0.400), improving draw recall substantially. However, this comes at the cost of reduced overall accuracy (57.89\%). Threshold tuning at various decision boundaries produces intermediate trade-offs. No configuration achieves Draw F1 above 0.40, and all draw-specific improvements reduce global accuracy.

This reveals a structural trade-off: improving minority-class sensitivity reduces global classification stability. Because Sim2Win is designed as a win-probability ranking system, the baseline CatBoost configuration is retained for deployment. However, users of the system should be explicitly informed that draw predictions are unreliable, and the system should not be used as a draw-detection tool without further architectural development.

\subsection{Explainability}

Interpretability was treated as a core design objective throughout the development of Sim2Win. Since the system is intended to support coaching and analytical decisions, model transparency is essential for trust and adoption.

\begin{figure*}[!t]
\centering
\includegraphics[width=1.02\textwidth]{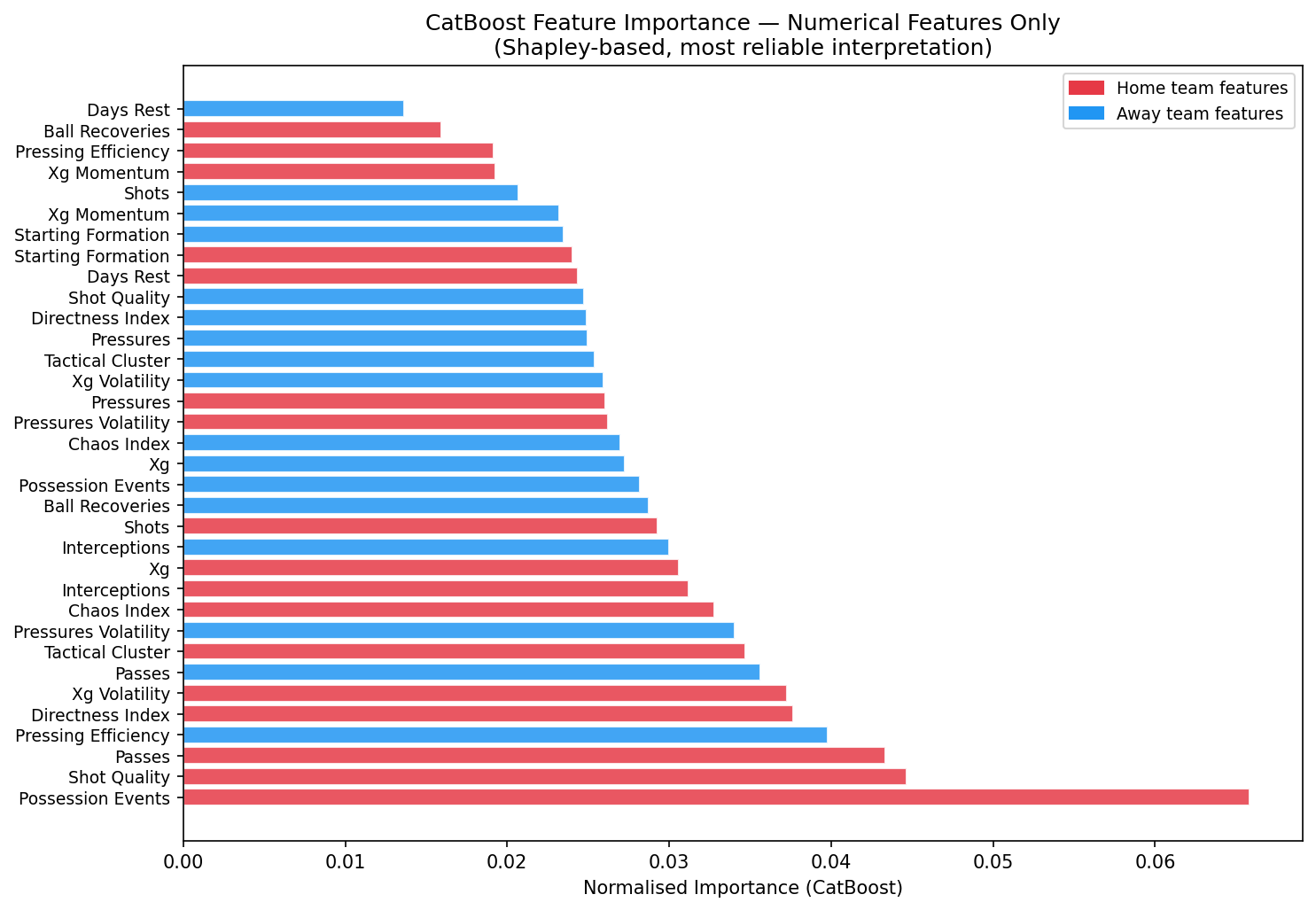}
\caption{CatBoost feature importance using Shapley-based importance values. The model relies primarily on possession events, shot quality, passes, pressing efficiency, and tactical volatility features. The balanced distribution between home and away tactical variables indicates that the system captures interaction dynamics between both teams rather than relying on single-team bias.}
\label{fig:feature_importance}
\end{figure*}

CatBoost feature importance was computed using Shapley-based importance values. Figure~\ref{fig:feature_importance} presents the normalized feature importance distribution across numerical tactical variables.

The most influential variables include possession events, shot quality, pass volume, pressing efficiency, tactical volatility indicators, and expected goals (xG). These features align with established football-domain expectations and validate the tactical relevance of the learned representations.

Possession events emerge as the strongest single feature, highlighting the importance of territorial control. The prominence of pressing-related variables suggests that defensive disruption and tactical consistency contain substantial predictive signal beyond traditional offensive metrics. The feature importance distribution remains balanced between home-team and away-team variables, indicating that the model captures interaction dynamics between both competing tactical systems rather than relying on a simplistic home-advantage proxy.

The relevance of tactical-cluster variables confirms that the playstyle abstraction layer provides contextual information to the model, consistent with the modest but consistent accuracy contribution observed in the ablation study.

\section{Discussion}
\label{sec:discussion}

The results demonstrate that rolling tactical behavioral representations maintain discriminative performance under cross-competition evaluation and outperform identity-based rating systems under distribution shift conditions. Beyond these core findings, several design choices and their limitations warrant careful discussion.

\subsection{Behavioral Modeling vs. Identity-Based Systems}

Conventional systems such as ELO, Pi-Rating, and GAP rely on persistent team identity. While effective where historical continuity exists, they become fragile when evaluating teams absent from the training distribution. Sim2Win models \emph{how} a team currently behaves rather than \emph{who} it is, using rolling event-based profiles that adapt naturally to evolving tactical structures.

The performance advantage over GAP (0.704 vs.\ 0.548 mean LOCO AUC) must be interpreted with the structural caveat noted in Section~\ref{sec:baselines}: GAP lacks team-level xG ratings for held-out competitions and reverts to an uninformative prior. In competitions where GAP retains historical data, the advantage narrows (e.g., World Cup: 0.680 vs.\ 0.618). A head-to-head comparison within known competitions, where both systems have full data, remains as future work and would provide a more complete picture of relative performance.

\subsection{Generalization Across Competitions}

The LOCO results reveal a meaningful generalization gradient. Strongest performance is achieved on La Liga and the Women's World Cup; weakest on AFCON and UEFA Euro. This variation is consistent with a training distribution dominated by European club football, which shares tactical characteristics with La Liga but diverges from African and international tournament football.

An important alternative explanation for this gradient deserves acknowledgment: Sim2Win's most influential features are home-team possession metrics, which partially proxy for structural home advantage. Competitions held entirely on neutral ground (World Cup, AFCON, Copa America) lack true home advantage, and these are precisely the competitions where Sim2Win performs more weakly. Whether the model's predictive signal derives primarily from tactical content or from home/away structural asymmetry cannot be fully disentangled from the current dataset. Controlled evaluation on neutral-ground matches exclusively would help isolate this effect and is recommended as a methodological check before deployment claims are generalized.

\subsection{The Prediction-Profiling Gap}

Sim2Win is framed as a decision-support system, but the current implementation provides playstyle classification and win-probability estimation rather than prescriptive tactical guidance. The system identifies which of eight tactical archetypes is predicted to maximize win probability against a given opponent profile, but does not specify how a coaching staff should modify formations, pressing triggers, or individual player assignments to transition toward that archetype.

Bridging this gap would require, at minimum, player-level capability data to assess the feasibility of tactical transitions and validation with domain experts to confirm that recommended clusters are operationally achievable. Until these elements are incorporated, Sim2Win should be understood as a tool that narrows the tactical search space for analysts rather than as an automated coaching instruction system.

\subsection{The Role of Rolling Tactical Profiles}

The rolling window architecture prevents temporal data leakage, smooths noisy single-match events, captures short-term tactical evolution, and reflects current form rather than historical legacy. The ablation study validates this design: both shorter and longer windows underperform the five-match configuration by margins ($>3.9\%$) that exceed the cross-validation standard deviation and are therefore robust conclusions.

Volatility and momentum features derived from these windows contribute meaningful predictive signal. xG volatility captures tactical consistency, while momentum reflects short-term trajectory shifts that are often relevant to match preparation.

\subsection{The Raw Features Ablation Result}

Raw per-match statistics outperform rolling aggregated features by $+7.10\%$ accuracy, but this configuration is not deployable: it accesses statistics from the match being predicted, constituting data leakage. This result is reported as a diagnostic upper bound that quantifies the information gap between pre-match behavioral estimates and actual in-match events. Reducing this gap through richer pre-match context, injury reports, squad rotation, weather conditions, referee tendencies is a primary direction for future work.

\subsection{Draw Prediction as a Structural Challenge}

Draw prediction failure is not unique to Sim2Win but reflects a broader structural challenge. Draws often emerge from balanced tactical interactions, conservative game-state management, or stochastic late-game dynamics that are difficult to infer from pre-match behavioral profiles. The current system's draw detection is insufficient for applications that require it. More advanced formulations such as ordinal classification or multi-stage probabilistic architectures may provide stronger solutions.

\subsection{Clustering and Tactical Diversity}

The $K=8$ choice overrides the mathematical optimum ($K \approx 5$) suggested by elbow and silhouette analysis. The rationale is that richer cluster granularity improves the practical usability of the interpretation interface. However, this choice is not validated through external expert annotation or inter-rater agreement on cluster labels, and the assignments should be treated as exploratory groupings rather than established tactical categories. Stability analysis across random seeds and expert validation are recommended before the cluster taxonomy is used in professional settings.

\subsection{Broader Implications for Sports Analytics}

The findings demonstrate the effectiveness of behavior-based representations under distribution shift, limited historical continuity, and dynamic tactical evolution. They also suggest that prediction-oriented systems benefit from interpretable components not because interpretability improves accuracy, but because it determines whether practitioners will trust and use the system's output. More broadly, this work contributes to the evolution of sports analytics from descriptive statistics toward decision-support systems that integrate prediction, behavioral profiling, and tactical reasoning.

\section{Limitations and Future Work}
\label{sec:limitations}

\subsection{Dataset Scale and Distributional Bias}

The dataset contains 1,411 team-match rows, which is small by modern machine learning standards for a multi-class prediction problem with a high-dimensional feature space. Several ablation effects ($\Delta\text{Acc} < 1\%$) fall within the cross-validation standard deviation and should be interpreted as directional rather than statistically confirmed. Competition-level imbalance with domestic European leagues dominating the training distribution likely contributes to weaker generalization on AFCON and international tournaments. Expanding to additional leagues and seasons is the primary dataset improvement needed.

\subsection{Home Advantage Confound}

The model's strongest feature is home-team possession events, and confusion matrices reveal systematic over-prediction of home wins. A portion of Sim2Win's predictive signal may therefore derive from the structural home advantage in football rather than from tactical content. This confound cannot be resolved with the current dataset, which contains no truly neutral-ground domestic matches. Evaluation on a neutral-ground-only subset (e.g., World Cup knockout rounds) is recommended to quantify this effect before deployment claims are generalized.

\subsection{Draw Prediction Failure}

The system predicts zero draws in at least one LOCO fold and achieves Draw F1 of 0.238 on the holdout set. This represents a structural failure for any use case requiring draw detection. No tested configuration eliminates this limitation without a corresponding reduction in overall accuracy. The system should not be used for draw prediction without further architectural development.

\subsection{Baseline Comparison Asymmetry}

The comparison against ELO, Pi-Rating, and GAP under
LOCO is conducted under distribution shift conditions:
rating systems receive team names and are warmed up
chronologically on training matches, but the LOCO protocol
evaluates them on competitions absent from their training
distribution. This comparison therefore measures robustness
under distribution shift rather than head-to-head predictive
performance in the intended deployment context of rating
systems, where historical continuity exists. The reported
21/21 AUC advantage should not be generalized as evidence
that Sim2Win outperforms identity-based systems in standard
deployment scenarios. A within-competition evaluation where
all systems have complete historical data remains as future work.

\subsection{Cluster Validation}

The $K=8$ cluster choice is not validated through expert annotation, inter-rater agreement, or seed stability analysis. The eight tactical archetypes should be treated as an exploratory interpretation layer rather than a validated taxonomy of football playing styles.

\subsection{Limited Contextual Representation}

The current feature set omits several important contextual variables: player injuries and suspensions, squad rotation and fatigue, coach changes, weather conditions, referee tendencies, and psychological effects. Their absence limits the realism of the prediction system in scenarios where these factors materially affect tactical execution.

Future work should investigate integration of richer pre-match contextual information while preserving the team-agnostic nature of the framework. Additionally, spatial-temporal representations using StatsBomb 360 freeze-frame data or full tracking datasets would allow modeling of positional organization and collective movement patterns that cannot be captured by aggregated event statistics.

\subsection{Future Research Directions}

Priority directions for future work include: (1) expanding the dataset with additional seasons and leagues to reduce distributional imbalance; (2) evaluating performance on neutral-ground matches exclusively to isolate the home advantage confound; (3) implementing a within-competition baseline comparison where rating systems have full historical data; (4) validating cluster assignments through expert annotation; (5) exploring ordinal classification or multi-stage architectures for draw prediction; and (6) investigating hybrid architectures combining behavioral representations with macro-level rating systems to balance short-term tactical state with long-term team strength.

\section{Conclusion}
\label{sec:conclusion}

This paper introduced Sim2Win, a team-agnostic, event-based match outcome prediction system with a playstyle-based interpretation layer for football. Unlike traditional prediction systems that rely primarily on persistent team identity or historical reputation, Sim2Win focuses on tactical behavior, representing teams through rolling event-based profiles that generalize across competitions without requiring identity information.

The framework combines StatsBomb event data, rolling five-match tactical profiles, interpretable feature engineering, K-Means playstyle clustering, CatBoost prediction, LOCO validation, SHAP-based explainability, and Gemini-generated coaching narratives within a unified prediction and profiling pipeline.

Experimental results demonstrate strong generalization capability. Across seven unseen competitions under the Leave-One-Competition-Out framework, Sim2Win achieves 55.4\% mean accuracy and 0.704 mean ROC-AUC, while maintaining performance superior to ELO, Pi-Rating, and GAP under distribution shift conditions. We note that this comparison is structurally asymmetric rating systems lack historical data for held-out teams and should be interpreted as evidence of robustness under distribution shift rather than general superiority.

The findings show that rolling tactical behavioral representations generalize more effectively than identity-based rating systems under cross-competition evaluation and capture meaningful football concepts such as possession control, shot quality, pressing efficiency, and tactical volatility. Key limitations draw prediction failure, home advantage confound, and unvalidated cluster assignments are acknowledged and represent concrete directions for future work.

Overall, Sim2Win provides a transferable, interpretable, and deployable foundation for AI-assisted football tactical profiling and contributes toward the evolution of sports analytics from descriptive prediction toward behavioral decision-support systems.

{\small
\bibliographystyle{ieeetr}
\bibliography{references}
}

\end{document}